\documentclass[twocolumn]{IEEEtran}    
\usepackage{cite}
\usepackage[switch,pagewise]{lineno}
\usepackage{url,subfigure}
\usepackage[hidelinks,hypertexnames=false]{hyperref}
\usepackage{booktabs}
\usepackage{multirow}
\usepackage{graphicx}
\usepackage{makecell}
\usepackage{ulem}
\usepackage{color}
\usepackage{tabularx}
\usepackage{float}

\usepackage{scalerel}
\usepackage{tikz}
\usetikzlibrary{svg.path}
\definecolor{orcidlogocol}{HTML}{A6CE39}
\tikzset{
  orcidlogo/.pic={
    \fill[orcidlogocol] svg{M256,128c0,70.7-57.3,128-128,128C57.3,256,0,198.7,0,128C0,57.3,57.3,0,128,0C198.7,0,256,57.3,256,128z};
    \fill[white] svg{M86.3,186.2H70.9V79.1h15.4v48.4V186.2z}
                 svg{M108.9,79.1h41.6c39.6,0,57,28.3,57,53.6c0,27.5-21.5,53.6-56.8,53.6h-41.8V79.1z M124.3,172.4h24.5c34.9,0,42.9-26.5,42.9-39.7c0-21.5-13.7-39.7-43.7-39.7h-23.7V172.4z}
                 svg{M88.7,56.8c0,5.5-4.5,10.1-10.1,10.1c-5.6,0-10.1-4.6-10.1-10.1c0-5.6,4.5-10.1,10.1-10.1C84.2,46.7,88.7,51.3,88.7,56.8z};
  }
}
\newcommand\orcidicon[1]{\href{https://orcid.org/#1}{\mbox{\scalerel*{
\begin{tikzpicture}[yscale=-1,transform shape]
\pic{orcidlogo};
\end{tikzpicture}
}{|}}}}
\usepackage{hyperref}
\hypersetup{
    colorlinks=true,
    linkcolor=blue,
    filecolor=black,      
    urlcolor=black,
    citecolor=purple
}
\usepackage{tikz-network}

\newcommand{\settablefont}{\fontsize{6.9}{11.8}\selectfont}
\newcommand\clr[1]{{\color{black}{#1}}}
\newcommand\cll[1]{{\color{black}{#1}}}
\newcommand\clv[1]{{\color{black}{#1}}}
\newcommand\clb[1]{{\color{black}{#1}}}
\begin{document}

\normalem
\title{Establishing Reality-Virtuality Interconnections in Urban Digital Twins for Superior Intelligent\\Road Inspection \clb{and Simulation}}


\author{
Yikang Zhang$^{\orcidicon{0009-0003-8840-392X}\,}$,
Chuang-Wei Liu$^{\orcidicon{0000-0003-0260-6236}\,}$,
Jiahang Li$^{\orcidicon{0009-0005-8379-249X}\,}$,
Yingbing Chen$^{\orcidicon{0000-0003-2004-9816}\,}$,\\
Jie Cheng$^{\orcidicon{0000-0002-1507-0074}\,}$,
Rui Fan$^{\orcidicon{0000-0003-2593-6596}\,}$,~\IEEEmembership{Senior Member,~IEEE}
\thanks{Yikang Zhang, Chuang-Wei Liu, Jiahang Li and Rui Fan are with the College of Electronics \& Information Engineering, Shanghai Research Institute for Intelligent Autonomous Systems, the State Key Laboratory of Intelligent Autonomous Systems, and Frontiers Science Center for Intelligent Autonomous Systems, Tongji University, Shanghai 201804, China. 
Yingbing Chen is with Individualized Interdisciplinary Program, Division of Emerging Interdisciplinary Areas, HKUST.
Jie Cheng is with the Department of Electronic and Computer Engineering, HKUST, Hong Kong SAR
(e-mails: 
\{yikangzhang, cwliu, lijiahang617\}@tongji.edu.cn,
\{ychengz, jchengai\}@connect.ust.hk,
rui.fan@ieee.org
).}
}
\maketitle	

\begin{abstract}
Road inspection is crucial for maintaining road's \cll{serviceability} and ensuring traffic safety, as road defects gradually develop and compromise functionality. Traditional inspection methods, which rely on manual evaluations, are labor-intensive, costly, and time-consuming.
While data-driven approaches are gaining traction, the scarcity and spatial sparsity of real-world road defects present significant challenges in acquiring high-quality datasets. Existing simulators designed to generate detailed synthetic driving scenes, however, lack models for road defects. Moreover, advanced driving tasks that involve interactions with road surfaces\cll{,} such as planning and control in defective areas\cll{,} remain underexplored.
To address these limitations, we propose a multi-modal sensor platform integrated with an \cll{u}rban \cll{d}igital \cll{t}win (UDT) system for intelligent road inspection. First, hierarchical road models are constructed from real-world driving data collected using vehicle-mounted sensors, \cll{resulting in} highly detailed representations of road defect structures and surface elevations. Next, digital road twins are generated to create simulation environments for comprehensive analysis and evaluation \cll{of algorithm's performance}. These scenarios are then imported into a simulator to facilitate both data acquisition and physical simulation.
Experimental results demonstrate that driving tasks, including perception and decision-making, benefit significantly from the high-fidelity road defect scenes generated by our system.
\end{abstract}


\section{Introduction} 
\label{sec:intro}

\IEEEPARstart{R}{oad} \cll{condition assessment is essential for ensuring optimal} vehicle dynamics and driving performance \cite{guo2018review}. \cll{Defects} such as cracks and potholes \cll{not only induce} vibrations \cll{but also} accelerate the wear of vehicle components \cite{fan2021rethinking}. Timely detection and repair of these defects are \cll{therefore crucial for ensuring} traffic safety \cite{fan2019road}. \cll{Nevertheless}, current road inspection methods face \cll{significant} challenges \cite{kim2014review}. Traditional manual visual inspection, \cll{carried out} by certified inspectors, is both labor-intensive and hazardous \cite{fan2019pothole}. \cll{It also causes substantial disruption} to traffic flow, making it impractical for large-scale road condition assessment \cite{wang2024two}. \cll{In recent years}, data-driven approaches have \cll{gained prominence}. By fully leveraging \cll{extensive} driving datasets, \cll{deep neural networks (DNNs)} can achieve outstanding performance \cite{zhao2023drmnet}. However, despite the unlimited data that can be collected, their effectiveness remains \cll{limited} by the availability of high-quality ground-truth annotations \cite{yin2024promoting}.

Recent technological advancements have \cll{enabled the development} of intelligent road inspection systems supported by \cll{u}rban \cll{d}igital \cll{t}wins (UDT).
A UDT system reconstructs the semantic and geospatial properties of urban entities, such as roads and buildings \cite{guo2024udtiri}, \cll{providing indispensable digital replicas for} real-time monitoring and simulation. However, existing simulators typically represent roads as 2D planar surfaces, \cll{unable to capture surface unevenness}. \cll{As road surfaces directly interact with vehicles and significantly influence driving safety, it is necessary to reconstruct their 3D structures using real-world measurements.} 

To address these challenges, we develop a UDT system that enables reality-virtuality interconnection (RVI), specifically tailored for intelligent road inspection. The system not only \cll{creates} digital twins of road entities, but also \cll{builds simulation environments containing road defects to support} the evaluation of perception and decision-making algorithms.
\cll{For perception tasks such as semantic scene parsing and 3D geometry reconstruction, it not only replicates real-world road environments but also generates novel scenarios by randomly combining road models to enhance generalization. 
Experiments show that semantic segmentation and stereo matching networks pre-trained on our synthetic data transfer effectively to real-world datasets.
For decision-making tasks, we explore a new paradigm for handling road defects. Instead of following conventional methods that treat defects as obstacles to be avoided entirely, our approach enables either safely bypassing or gliding over certain defects without full detours, leveraging detailed road surface geometry from the UDT. By incorporating tire-level collision detection rather than vehicle-body bounding box checks, the system supports more flexible and efficient obstacle avoidance strategies.}

\begin{figure}[!t]
    \centering
    \includegraphics[width=0.45\textwidth]{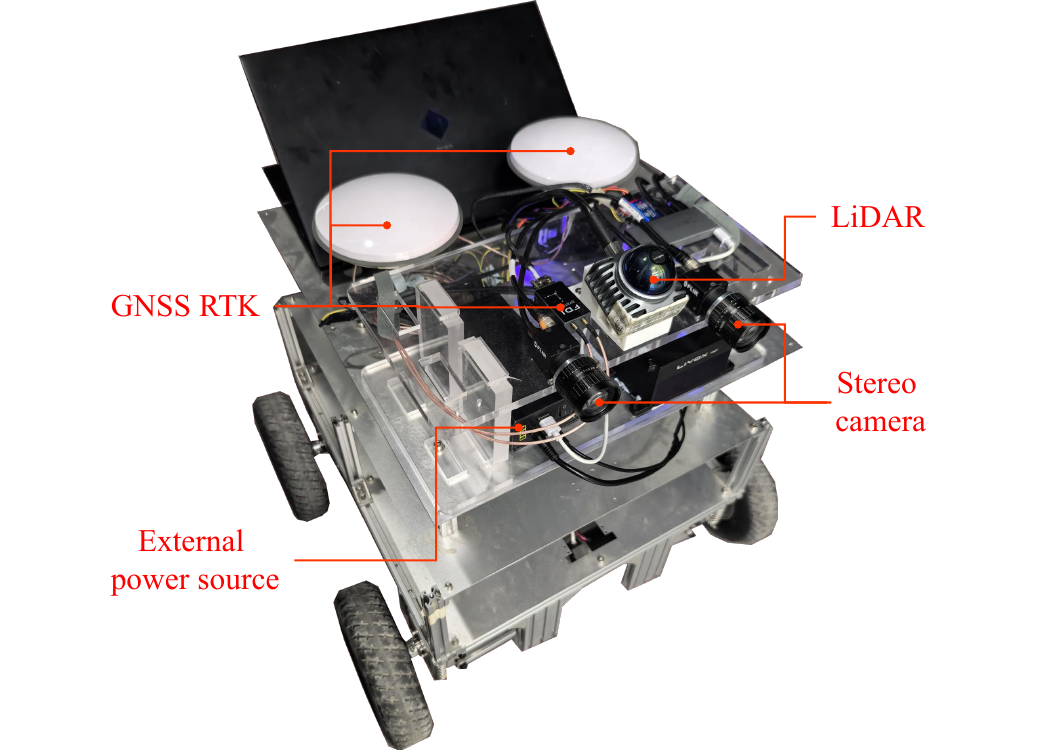}
    \caption{ The sensor setup includes a LiDAR with an integrated IMU, a stereo camera pair, and a GNSS RTK module, all powered by an external battery. A high-performance laptop processes the acquired data. 
    }
    \label{fig.sensor_platform}
\end{figure}

The contributions of this study span five key aspects: equipment, algorithm, simulator, dataset, and benchmark. As illustrated in Fig. \ref{fig.sensor_platform}, we develop a portable, multi-sensor experimental setup, equipped with a Livox Mid-360 LiDAR, two FLIR BFS-U3-31S4C cameras, and a DETA100D4G GNSS RTK module. This setup can be mounted on any vehicle, enabling multi-modal, high-quality road data collection in real-world scenarios.  
To model the interconnection between reality and virtuality for urban entities, we develop two key modules: a hierarchical road model creator and a digital road twin generator.
The former module extracts road defect structures and undamaged surface elevations from real-world sensor data. 
\cll{Real-world road entities across different scales, from coarse-grained road surfaces that shape the overall driving environment to fine-grained defects that directly affect vehicle stability and comfort,} are segmented and integrated into unified road surface mesh models, replacing the simplified planar road assets for more realistic road scenes.
The latter module creates high-fidelity scenes for synthetic data collection and physical simulation. 
Given the challenges inherent in large-scale road surface and sporadically occurred road defects in \cll{the} real world, our system enables seamless integration of any defect model with any surface model smoothly, even if the models originate from different locations.
For perception tasks, we create a comprehensive dataset containing semantic and instance-level annotations, along with ground truth data for depth, event, and surface normals, to benchmark state-of-the-art (SoTA) DNNs for intelligent road inspection. 
\cll{For decision-making tasks, we design an experimental framework that treats road defects as negative obstacles, enabling the evaluation of diverse trajectory planning and speed control algorithms. The framework provides physical simulation for both defect-avoidance and traversal strategies that optimize driving comfort, incorporating wheel-level collision constraints for more precise decisions.}

In a nutshell, our main contributions are as follows:
    \begin{itemize}
    \item We design a UDT system for road inspection, including a portable multi-modal sensor equipment for real-world data collection, and \cll{simulation environments containing road defects} for data synthesis and physical simulation.
    \item We propose a pipeline to autonomously construct hierarchical road models and generate digital road twins, \cll{which both replicate high-fidelity real-world environments and enable the creation of novel simulation scenarios.}    
    \item \cll{We provide a comprehensive benchmark to evaluate perception and decision-making algorithms leveraging our urban digital twins, demonstrating their effectiveness for downstream driving tasks.}
\end{itemize}

The remainder of this article is structured as follows:
Sect. \ref{sec.relatedwork} reviews relevant research works. 
Sect. \ref{sec.udt_system} \cll{introduces the UDT system} for intelligent road inspection.
Sect. \ref{sec.experiment} presents the simulation setup and experimental results.
Finally, Sect. \ref{sec.discussion} \cll{discusses the limitations and potential directions for future research, while Sect. \ref{sec.conclusion} concludes the study.}

\section{Related Works}
\label{sec.relatedwork}

\subsection{Intelligent Road Inspection}
Traditional road inspection is typically conducted by structural engineers or certified inspectors, a process that is often subjective, inefficient, and sometimes hazardous \cite{kim2014review}.
To address these challenges, researchers have developed intelligent road inspection systems capable of automatically collecting data and identifying road defects \cite{ma2022computer}. 
Among the most direct effects of road defects is vibration. Measurement systems analyze vehicle dynamics as the vehicle traverses road defects \cite{dozsa2022road}. While these systems provide high precision, they are limited to detecting defects within the tire tracks.
Other approaches mount visual sensors on the rear of the vehicle with a downward-facing view, maintaining a known distance from the road surface, including single cameras \cite{prasanna2014automated}, stereo camera pairs \cite{karunasekera2019energy}\cll{,} and laser scanners \cite{han2012enhanced}. 
However, such specialized sensor configurations are restricted to dedicated inspection vehicles, limiting scalability and widespread applicability.
In contrast, crowd-sourcing approaches collect large-scale data from daily driving using forward-facing sensors, enabling dual functionality for both driving perception and road inspection.
For instance, LiDAR-based methods collect data for road unevenness \cite{zhao2023hierarchical}, while stereo camera-based approaches reconstruct 3D road surfaces \cite{oniga2009processing}. Smartphone applications have also emerged as a practical data source.

Numerous algorithms have been developed for road defect detection.
Classical 2D image processing methods have been widely explored, extracting damaged road areas from segmented foregrounds based on geometric and textural assumptions \cite{fan2021rethinking}. 
However, such methods are sensitive to environmental factors that violate the underlying assumptions.
3D point cloud methods have also been employed to detect road irregularities. Stereo vision-based approaches reconstruct dense 3D road point clouds and interpolate them into planar or quadratic surfaces. LiDAR-based methods are utilized for road roughness perception \cite{zhao2024roadbev}, while road patches traversed by a vehicle’s tires are segmented to identify irregularities such as bumps and potholes \cite{RSRD}.
With the rise of deep learning, data-driven approaches have become the dominant techniques \cite{feng2022mafnet}.
However, the performance heavily depends on the dataset quality. Due to the rarity and sparse distribution of road defects, capturing sufficient annotated data for training remains a significant challenge.

\subsection{Urban Digital Twins}
Digital \cll{t}win, originally developed for cyber-physical integration in factories, connects real and virtual products through advanced data communication technologies \cite{lu2020communication}.
Leveraging advantages such as real-time monitoring, fast simulation and troubleshooting, digital twin technology applied to urban entities integrates online sensor data and creates virtual models to address road anomalies \cite{wang2024applications}. These capabilities are exemplified by applications such as intelligent road inspection \cite{fan2022urban}.
For example, \cite{cao2022pavement} proposed a pavement crack segmentation approach based on 3D edge detection within digital twins. 
Recent studies have also achieved large-scale reconstruction of explicit road surface meshes, recovering both geometry and texture \cite{mei2024rome, wu2024emie}. However, most research treats digital twins of road defects and surfaces as individual reconstruction tasks.
In contrast, our work integrates defective road entities at the scene level, creating comprehensive simulation environments specifically tailored for autonomous driving applications.

\section{Urban Digital Twin System}
\label{sec.udt_system}

\begin{figure}[!t]
    \centering
    \includegraphics[width=0.485\textwidth]{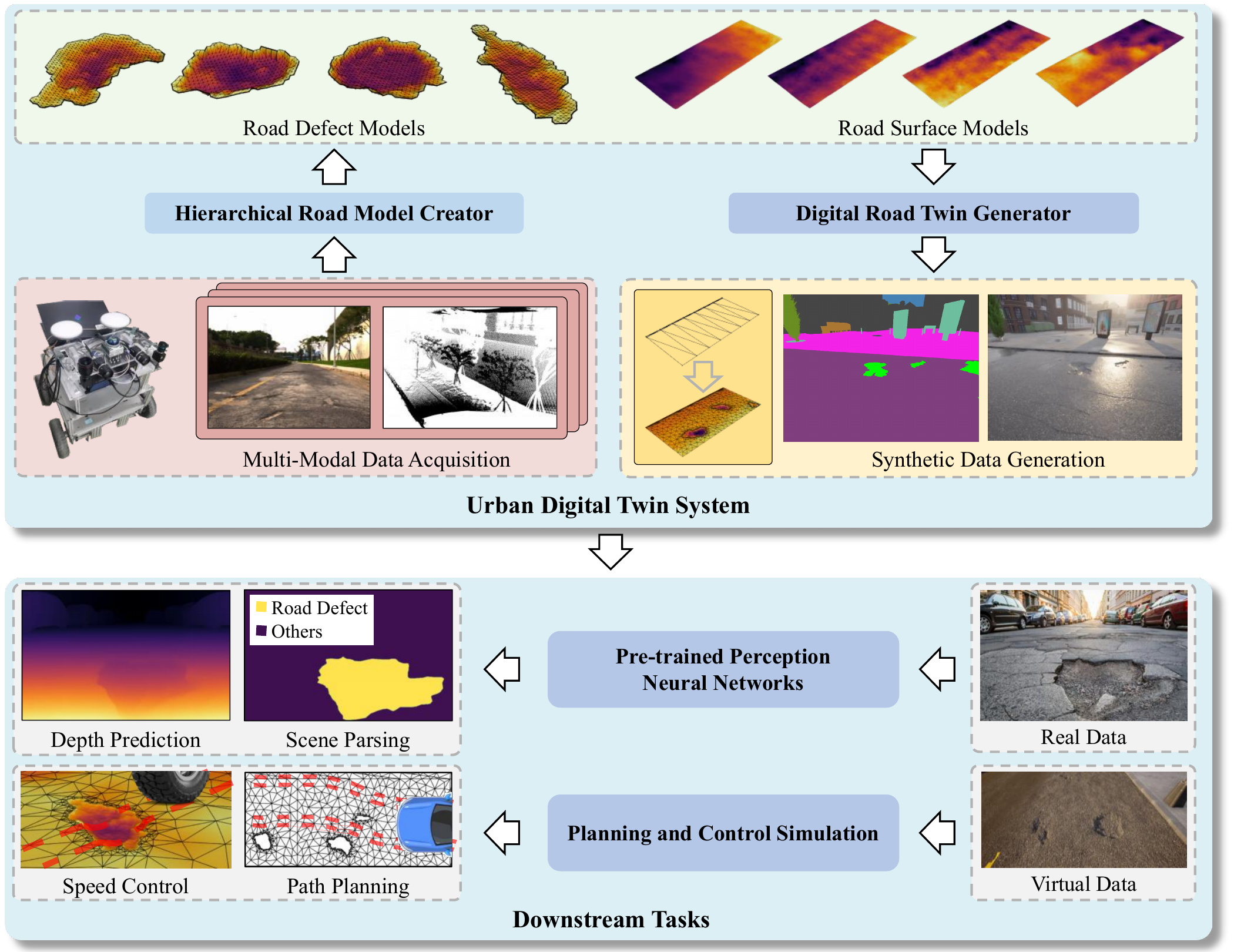}
    \caption{The UDT system is composed of two components. The hierarchical road model creator autonomously reconstructs 3D defect models and non-defective road surfaces from the physical world, while the digital road twin generator produces virtual entities for a diverse, well-annotated environment. } 
    \label{fig.overview}
\end{figure}

\begin{figure}[!t]
    \centering
    \includegraphics[width=0.485\textwidth]{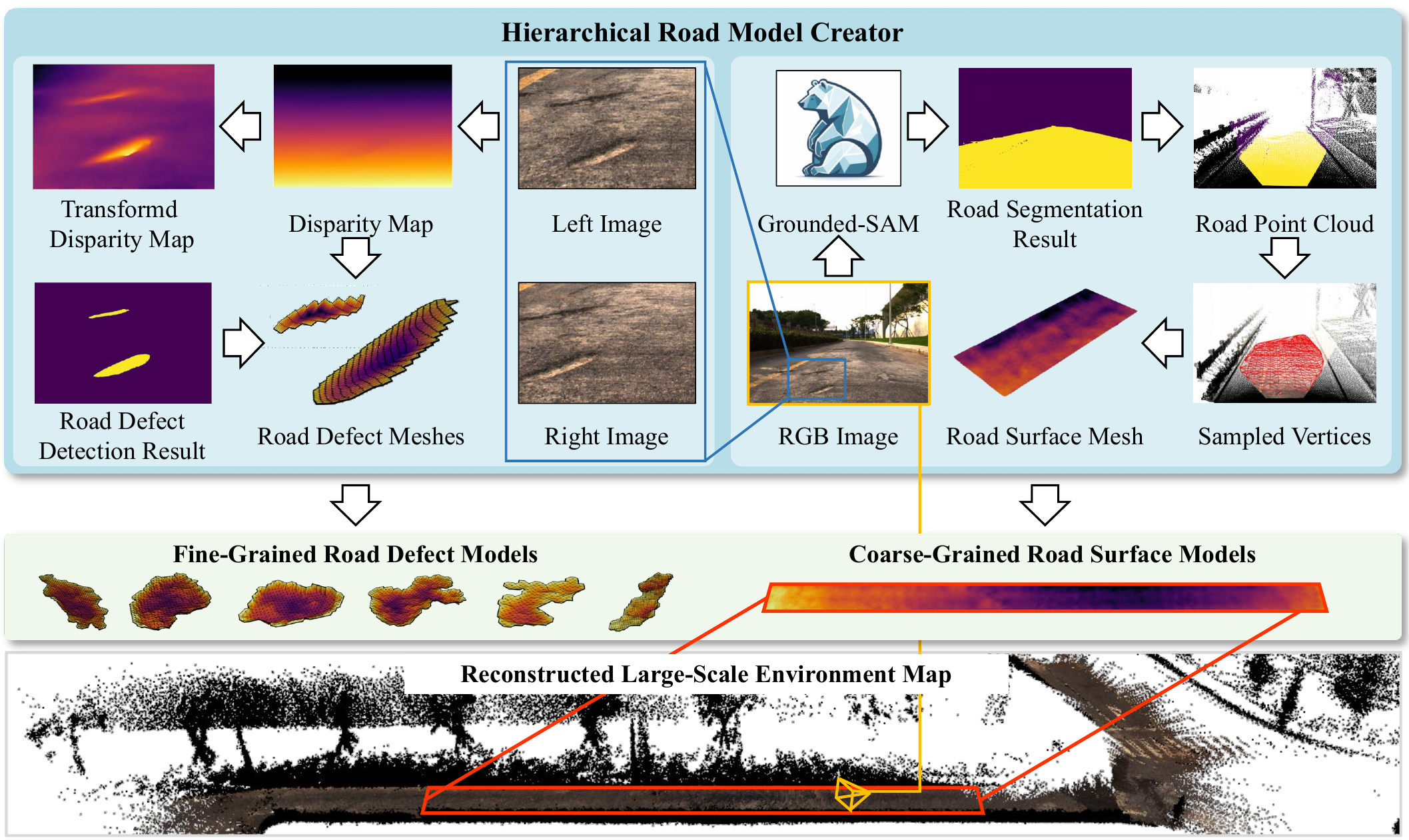}
    \caption{ Our hierarchical road model creator consists of two reconstruction streams. In the coarse-grained stream, semantic annotations are generated by Grounded-SAM, which filters LiDAR points for road surface mesh generation. 
    The fine-grained stream reconstructs road defect models using transformed disparity maps derived from stereo images.
    The reconstructed results are stored in a library for future use.
    }
    \label{fig.hierarchical_road_model_creator}
\end{figure}

\cll{\subsection{Hierarchical Road Model Creator}}
\label{sec.hrmc}
The hierarchical road model creator consists of two reconstruction streams, one for road defects and the other for non-defective surfaces, as illustrated in Fig.~\ref{fig.hierarchical_road_model_creator}.
The coarse-grained stream, focused on the reconstruction of non-defective road surfaces, follows a straightforward process.
Semantic masks are generated from camera images using Grounded-SAM \cite{groundedsam}, with prompts such as ``road" or ``pavement".
LiDAR points are segmented based on these masks and fused across multiple frames. To achieve coordinate alignment, a horizontal ground plane is fitted to points from multiple road segments, assuming local elevation variations while maintaining overall flatness.
A 3D downsampling operation is then applied to refine the vertices before meshing, accounting for density variations in the point clouds caused by redundant or missing observation views.
Given the simple geometry of non-defective road surfaces, elevation can be decoupled from the planar coordinates. The mesh model is reconstructed by directly triangulating the 2D projection of vertices onto the horizontal plane. Vertex elevations are subsequently restored from the original point cloud, ensuring that the reconstructed surface accurately reflects real-world topography.

\begin{figure}[!t]
    \centering
    \includegraphics[width=0.49\textwidth]{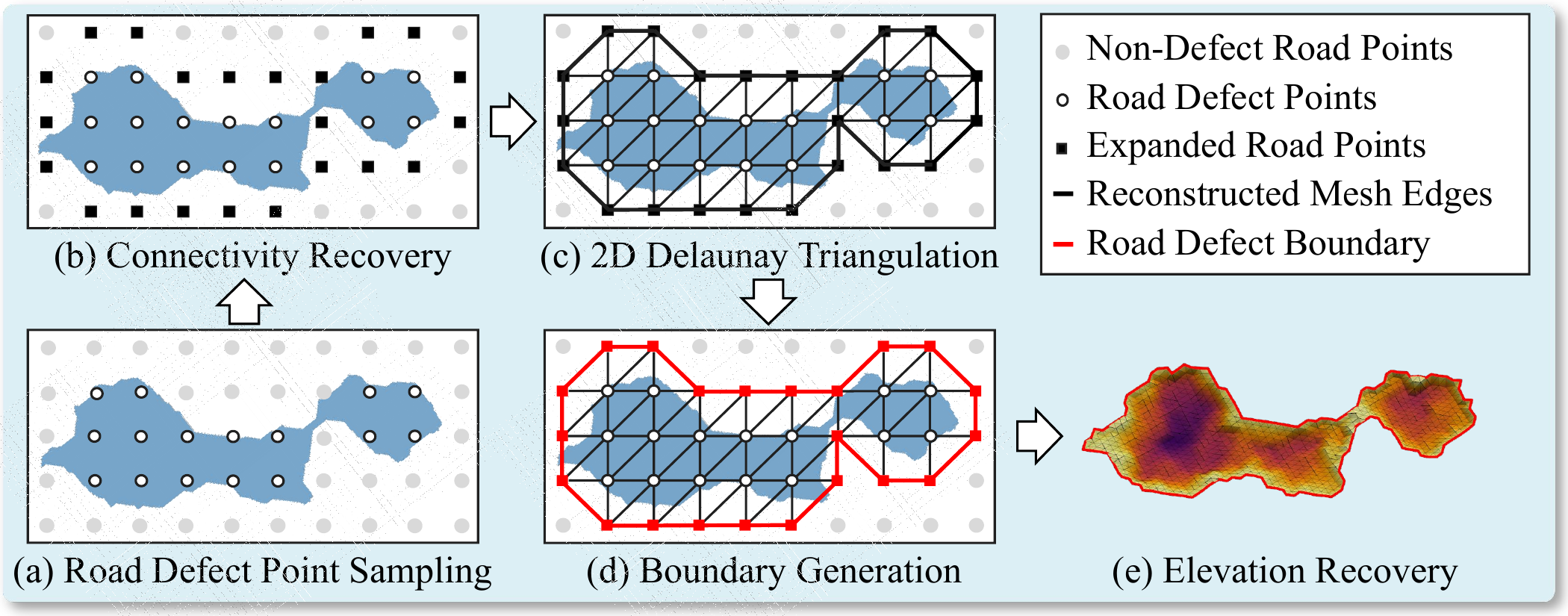}
    \caption{Example of the road defect reconstruction process. Points are iteratively sampled until the entire road defect structure is covered, ensuring topological connectivity. After triangulation, the boundary vertices are stored in a list with pseudo-heights for further structural alignment in the digital road twin generator.
    }
    \label{fig.fine_grained_defect_reconstruction}
\end{figure}

The stream for fine-grained road defects, unlike the non-defective road surfaces, requires dense geometric details.
Although many neural networks are proposed for road defect detection from monocular images, semantic cues at the boundaries between defective and non-defective surfaces are often ambiguous, leading to artifacts such as isolated pixels or incomplete segmentation. Therefore, defect regions are extracted based on geometric cues, performed by transformed disparity maps derived from stereo images, which highlight spatial deviations from the planar patches of the road surface.
Road defect instances are identified as connected components within the parsed scene. Instances with regular shapes, as shown in Fig. \ref{fig.hierarchical_road_model_creator}, can be reconstructed directly through grid-based sampling, similar to the first stream.
However, defects with irregular shapes, as illustrated in Fig. \ref{fig.fine_grained_defect_reconstruction}, may introduce structural discontinuities, even when pixel connectivity is preserved.
To address this issue, the sampled point set is expanded iteratively until the entire defect region is covered.
Expanded points, belonging to the defect but lying on the non-defective region, are shared between adjacent meshes.
If reconstructed separately, these duplicate boundary points could cause elevation mismatches, resulting in gaps or overlaps. 
Therefore, the boundary of the defect mesh is extracted and assigned a pseudo-height, which becomes active only when assembled with neighboring road surface meshes.
A boundary edge is defined as an edge connected to only one face in the mesh, indicating that it forms the outer boundary or ``hole'' of the surface.
All faces in the road defect model are traversed to identify such edges. 
Since boundary points are guaranteed to lie on the non-defective road surface, they can be used to fit a transformation matrix that aligns the defect model with the ground plane, ensuring that the pseudo-height is set to zero.

The outputs from the two reconstruction streams are compiled into a road model library. This design, with varying levels of granularity, ensures that the road meshes remain memory-efficient while retaining sufficient details.

\subsection{Digital Road Twin Generator}
\label{subsec.scene_generation}
\begin{figure}[!t]
    \centering
    \includegraphics[width=0.49\textwidth]{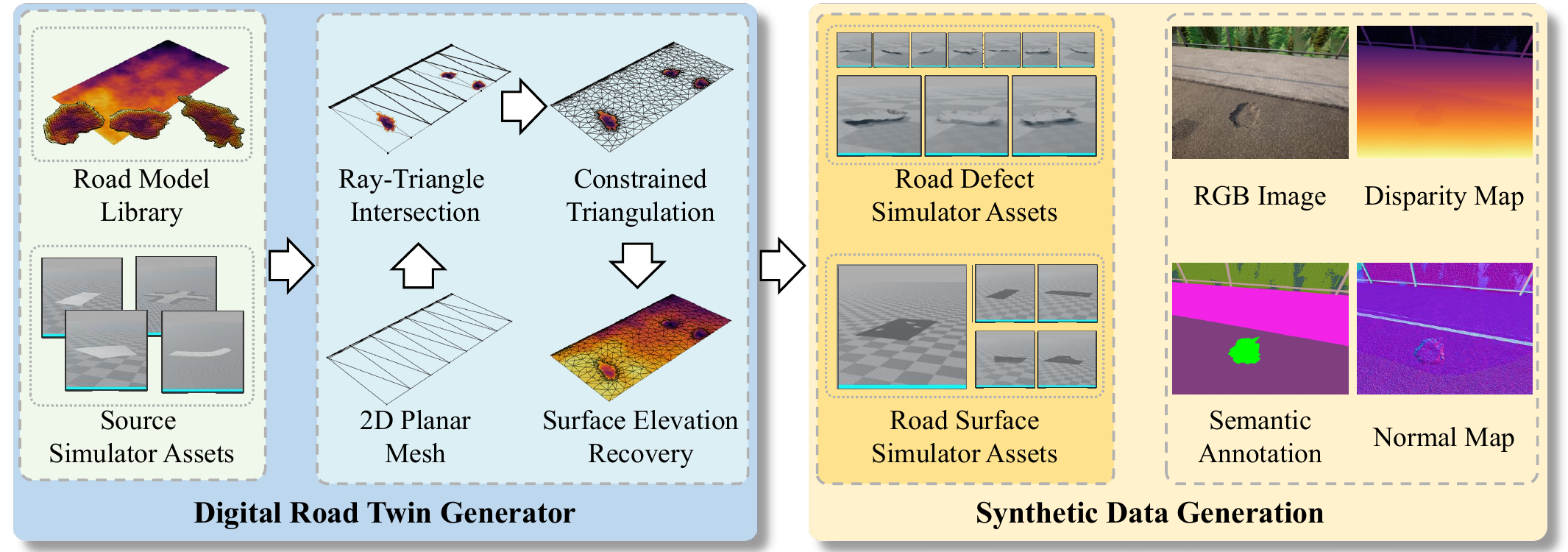}
    \caption{The Digital Road Twin Generator integrates simulator assets with reconstructed road models. Road defect models are projected onto the 2D road mesh to eliminate intersecting faces. The 2D structure is reorganized using constrained Delaunay triangulation, preserving model boundaries. For compatibility with the simulator, the generator's output is disassembled into individual road surface and defect asset groups, providing semantic references for the rendering pipeline.
}
    \label{fig.digital_road_twin_generator}
\end{figure}

Although road scenes can be accurately replicated, road defects are relatively rare in the physical world. To generate more diverse scenes, over-simplified road models from simulators must be updated with our road defect models, as illustrated in Fig.~\ref{fig.digital_road_twin_generator}.
First, the original planar road segment assets are exported from the simulator 
 to maintain road topology for the scene. The generated road defect models are then sampled with random poses and scales for integration. When merging the mesh models, intersecting triangular faces must be removed. However, because road segments are not always convex, simply removing all faces and re-meshing could compromise their structural integrity.
Therefore, defect models are projected vertically onto the road surface to accurately identify and remove only the intersecting triangles.
The Möller-Trumbore algorithm \cite{MollerTrumbore}\cll{, typically used to detect such ray-triangle intersections, is expressed as follows:}
\begin{align} \label{eq.ray_face}
    \boldsymbol{r}(t) &= \boldsymbol{o} + t\boldsymbol{d} = (1-u-v)\boldsymbol{p_0} + u\boldsymbol{p_1} + v\boldsymbol{p_2},\\
    s.t.&\begin{cases}
        0 \leq u \leq 1\cll{,} \\
        0 \leq v \leq 1\cll{,} \\
        0 \leq 1-u-v \leq 1\cll{,} 
    \end{cases}
    \end{align}
where ray $\boldsymbol{r}$ originates from $\boldsymbol{o}$ in direction $\boldsymbol{d}$ with length $t$, and the triangle is determined by points $\boldsymbol{p_0}$, $\boldsymbol{p_1}$, \cll{and} $\boldsymbol{p_2}$. Rewriting (\ref{eq.ray_face}) leads to \cll{the following expression}:
\begin{align}
    \boldsymbol{o} -\boldsymbol{p_0} &= (\boldsymbol{p_1}-\boldsymbol{p_0})u + (\boldsymbol{p_2} - \boldsymbol{p_0})v - t\boldsymbol{d},
\end{align}
which can be solved using Cramer's Rule, where
\begin{equation}
\begin{aligned}
    t &= \frac{(\boldsymbol{o} -\boldsymbol{p_0}) \times (\boldsymbol{p_1}-\boldsymbol{p_0})\cdot(\boldsymbol{p_2} - \boldsymbol{p_0})}{(\boldsymbol{o} -\boldsymbol{p_0}) \times (\boldsymbol{p_2}-\boldsymbol{p_0}) \cdot (\boldsymbol{p_1}-\boldsymbol{p_0}) },
\end{aligned} 
\end{equation}
with similar solutions for $u$ and $v$.
With all intersecting faces removed, the road defect models are positioned within the holes of the planar road mesh. 
The planar road polygons are then re-organized, typically through triangulation, which iteratively divides edges, samples vertices, and generates faces.
However, triangle mismatches may occur at the boundaries if adjacent meshes are reconstructed separately. 
For better realism and diversity, real-world road surface elevations are restored from the library.
The height of each vertex is efficiently queried through a \cll{\textit{k}-d} Tree built from the surface model. 
The final integrated 3D mesh is then split into individual assets, as required by the simulator, to provide semantic references.

Leveraging the hierarchical road model creator and the digital road twin generator described above, our UDT system bridges reality-virtuality interconnections to generate diverse, well-annotated simulation scenes featuring road defects.
The virtual environment not only provides synthetic data to enhance perception networks that depend on comprehensive road surface data, but also enables physical simulation for decision-making tasks that interact directly with road surfaces.
In the following section, we demonstrate these advancements by presenting perception tasks trained on synthetic data and decision-making tasks performed within our virtual environment.

\section{Experimental Results} 
\label{sec.experiment}

\subsection{\clr{Experimantal Setups}}
\clr{The sensor setup (Fig. \ref{fig.sensor_platform}) is mounted on a remotely-controlled vehicle for continuous data collection over campus roads.} 
\clb{Images and LiDAR point clouds are captured at 10Hz, while coarse GNSS coordinates are recorded to enable defect localization and support large-scale maintenance in the future. The GNSS module also outputs signals for hardware-level synchronization.}
\clr{The sensor poses are estimated using FAST-LIO \cite{fastlio}, which builds and updates a coarse-grained large-scale 3D road point cloud. The points belonging to drivable areas are filtered by Grounded-SAM, while stereo matching is applied to generate dense 3D reconstructions of detected defects \cite{fan2019pothole}. 
}
Based on sensor data collected from real-world defected roads, our hierarchical road model creator generates a library containing 53 distinct road defect models. 
These models are integrated into scenes in the simulator, replacing the original planar road surfaces.
For example, roads in CARLA Town01 are replaced by 646 synthetic road segments and 1,000 defect instances using our digital road twin generator.
\clr{In the simulation, an autopiloted vehicle navigates through road defect scenes populated with randomly placed surrounding vehicles and pedestrians. }
\clv{To mitigate the sim-to-real gap in the captured data, we introduce controlled randomness into the data acquisition pipeline. Specifically, sensor poses are sampled with random viewing angles rather than fixed trajectories, reducing view-dependent bias.}
\clr{Road surfaces and defects are rendered with diverse materials, including asphalt, cobblestones, and cement. Dynamic weather conditions are continuously simulated to enhance variability.
To demonstrate the effectiveness of our synthetic data, we selected models from MMSegmentation \footnote{\cll{MMSegmentation: \url{https://github.com/open-mmlab/mmsegmentation}}} with diverse backbones to ensure architectural diversity. For stereo matching, we employed representative state-of-the-art general-purpose models to provide a fair and rigorous baseline.
}

\subsection{Perception Tasks}
\subsubsection{Semantic Segmentation}

\begin{figure*}[!htbp]
    \centering
    \includegraphics[width=0.99\textwidth]{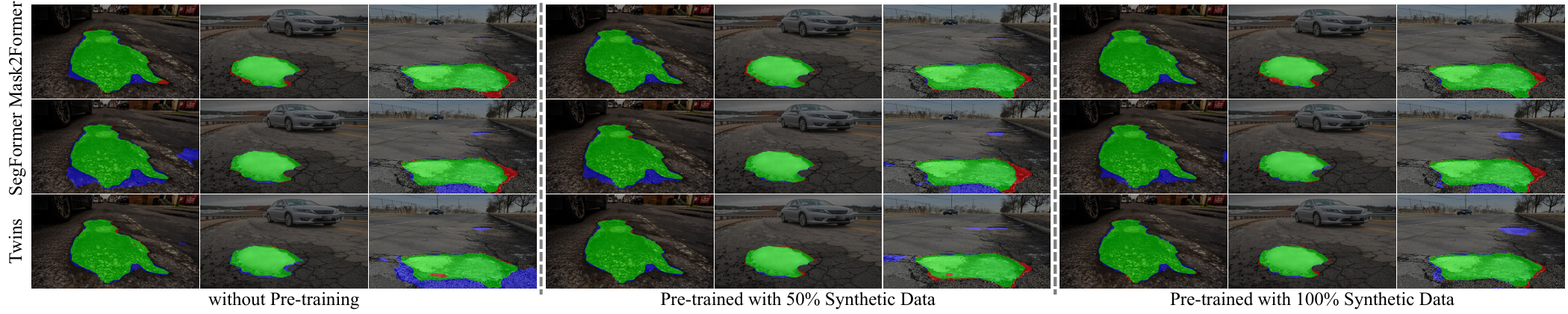}
    \caption{ Qualitative experimental results of semantic segmentation based on RGB images. 
    The green areas in the image represent true-positive predictions, the blue areas represent false-positive predictions, and the red areas represent false-negative predictions.
    }
    \label{fig.single_modal}
\end{figure*}

We first evaluate single-modal semantic segmentation networks using real-world RGB images. UDTIRI \cite{guo2024udtiri}, a real-world road defect dataset containing 1,000 semantically annotated images, is used for evaluation. 
All networks are pre-trained for 40,000 epochs by controlled ratios of synthetic data, followed by fine-tuning on 600 images from the real-world dataset. 
Network configurations follow the default settings of the MMSegmentation benchmark. \clr{To ensure fair comparisons, all hyperparameters are kept identical across experiments, with the only difference being the data for pretraining.}
Qualitative and quantitative results are presented in Fig. \ref{fig.single_modal} and Table \ref{tab:single_modal}, respectively. It is evident that pre-training with our synthetic data significantly enhances segmentation accuracy, providing more precise boundaries and reducing the occurrence of false-positive regions. The results show that synthetic data positively impacts image-based road defect detection, increasing mIoU by up to 3\%. 
This improvement alleviates the challenge posed by the limited availability of annotated road defect data for \cll{real-world} inspection models.

\begin{table}[t!]
\caption{mIoU(\%) $\uparrow$ of single-modal semantic segmentation results with different ratio of synthetic data}
\label{tab:single_modal} 
\centering
\settablefont
\renewcommand\arraystretch{0.8}{
\setlength{\tabcolsep}{7mm}{

\begin{tabular}{@{}l| c c c}

\toprule
\multirow{2}*{Method} &\multicolumn{3}{c}{Synthetic/Real data ratio} \\
\cline{2-4}
\multicolumn{1}{l|}{} & \multicolumn{1}{c}{0\%} & \multicolumn{1}{c}{50\%}  & \multicolumn{1}{c}{ 100 \%}  \\
\hline
\hline
Mask2Former & \multicolumn{1}{c}{77.08}    & \multicolumn{1}{c}{79.79} & \textbf{80.07}  \\ 
PSPNet     & \multicolumn{1}{c}{71.42}    & \multicolumn{1}{c}{71.44} & \textbf{74.29}  \\ 
SegFormer  & \multicolumn{1}{c}{75.16}    & \multicolumn{1}{c}{75.96} & \textbf{76.94}  \\ 
SegNext    & \multicolumn{1}{c}{79.68}    & \multicolumn{1}{c}{79.84} & \textbf{80.63}  \\ 
Twins      & \multicolumn{1}{c}{78.40}    & \multicolumn{1}{c}{\textbf{79.10}} & 79.06  \\ 
UperNet       & \multicolumn{1}{c}{70.97}    & \multicolumn{1}{c}{72.54} & \textbf{73.89}  \\ 
\bottomrule
\end{tabular}
}
}
\end{table}

\begin{table}[t!]
\caption{mIoU(\%) $\uparrow$ of \clr{multi-modal} semantic segmentation results. }
\label{tab:multi_modal} 
\centering
\settablefont
\setlength{\tabcolsep}{2mm}{
\begin{tabular}{@{}l| c c c c }
\toprule

Method & \multicolumn{1}{c}{RGB only}  & RGB+Depth  & RGB+Normal & RGB+Event  \\
\hline
\hline
RTFNet         & 67.5    &\textbf{74.6}  & 69.2 & 63.6\\ 
FuseNet  & 66.6    & \textbf{78.7} & 53.9 & 58.5  \\ 
MFNet             & 72.0    & 80.1 & \textbf{90.6} & 72.3 \\ 
SNE-RoadSeg      & 67.4    & 78.8 & \textbf{94.5} & 73.5  \\ 
RoadFormer   & 90.3   & 94.8 & \textbf{96.1} & 90.6  \\ 
\bottomrule
\end{tabular}
}
\end{table}

\begin{figure}[!hbtp]
    \centering
    \includegraphics[width=0.485\textwidth]{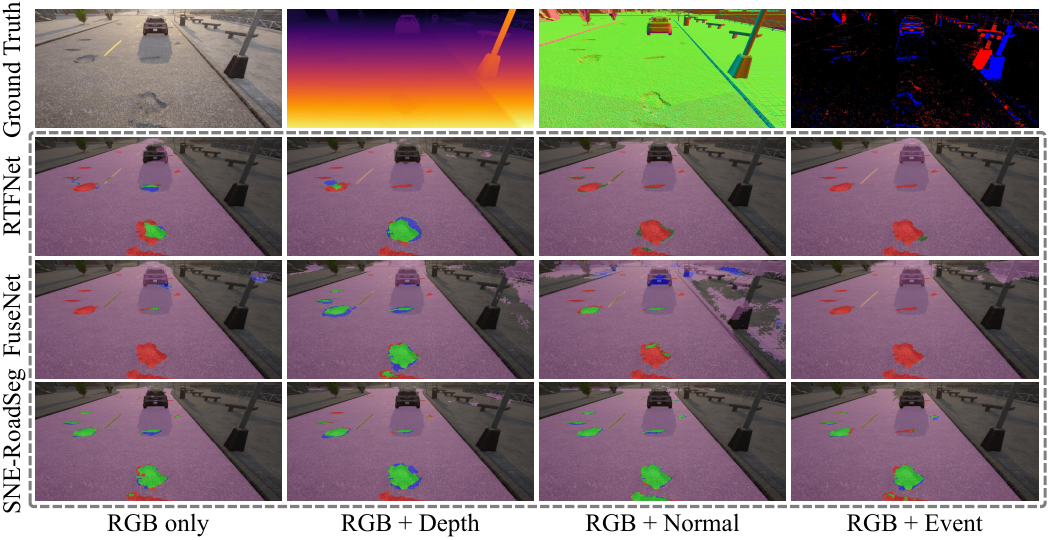}
    \caption{ Qualitative comparison between SoTA multi-modal networks on the synthetic data. The predicted drivable area is represented in purple. True-positive, false-positive, and false-negative classifications of road defects are shown in green, blue, and red, respectively.
    }
    \label{fig.multi_modal}
\end{figure}

\clb{Single-modal prediction results are often vulnerable to varying weather and illumination. In contrast, multi-modal fusion is widely regarded as a solution to enhance perception robustness by utilizing auxiliary information sources. However, capturing these modalities remains challenging. Leveraging the G-buffer in the rendering pipeline, we are able to evaluate auxiliary modalities for road inspection, focusing on image segmentation for drivable areas, road defects, and backgrounds, as shown in Fig. \ref{fig.multi_modal}}.
As shown in Table \ref{tab:multi_modal}, each additional modality is fused with RGB images in an RGB+X format. RGB+Depth provides the most generalized results, outperforming those trained using only RGB images by 4.5\% to 12.1\%. It also generates the most accurate true-positive predictions, demonstrating robust performance across all cases. This result is intuitive, as depth images directly capture the surface structure.
Normal maps, which are highly sensitive to object surfaces, improve performance by 1.7\% to 27.1\% across most networks, except for FuseNet. Although useful in many cases, this sensitivity also introduces noise, leading to convergence issues.
Simple fusion of event images across network channels is unstable. Although event images improve SNE-RoadSeg by 6.1\%, they degrade FuseNet's performance by 8.1\%. This instability is likely due to event data being influenced by the vehicle's driving speed, whereas RGB images assume static objects during exposure. To effectively utilize this modality, additional speed priors are required.

\subsubsection{Stereo Matching}

For stereo matching evaluation, we compare five SoTA networks to demonstrate the improvements enabled by our system. Similar to the segmentation experiments, \clr{all hyperparameters are kept fixed according to the released default settings, with the only difference being whether the models are pre-trained on synthetic UDT data}. 
Unlike individual road surface images that are \cll{available from crowd-sourcing applications,} stereo images of road surface defects are exceedingly rare.
To address this limitation, we fine-tune the networks using the KITTI dataset \cite{kitti}, a widely used driving dataset \cll{but does not contain} road defects. 
\cll{Experiments are conducted on both synthetic data with ground truth annotations and the real-world Stereo-Road dataset \cite{fan2019pothole} to evaluate zero-shot performance.}
Several metrics are selected for evaluation, including percentage of error pixels (PEP), which represents the percentage of incorrect disparities with respect to a tolerance of $\delta$ pixels, and end-point error (EPE), which measures the average disparity estimation error. 
\clb{Additionally, mean squared error (MSE), structural similarity index (SSIM), and peak signal-to-noise ratio (PSNR) are compared between the predicted and ground truth disparity maps.}
The loss function, learning rate, and optimizer settings are consistent with those reported in the corresponding publications.
Table \ref{tab:stereo_matching_on_synthetic} presents the quantitative results tested on synthetic data. Nearly all networks pre-trained with our synthetic data demonstrate performance improvements. EPE is improved across all networks, with reductions ranging from 0.60 to 3.18.
These results confirm that our UDT system significantly enhances model performance for stereo matching tasks.

\begin{table*}[!htbp]
\caption{
\clr{Comparisons of SoTA Stereo Matching Networks with and without our proposed synthetic data. }
}
\label{tab:stereo_matching_on_synthetic} 
\centering
\settablefont
\renewcommand\arraystretch{0.9}{
\setlength{\tabcolsep}{4mm}{

\begin{tabular}{l cccccc | ccc}
\toprule

\multicolumn{1}{l|}{\multirow{3}*{Method}} & \multicolumn{6}{c|}{Evaluation on Synthetic UDT Data} & \multicolumn{3}{c}{\clr{Evaluation on Stereo-Road Dataset (Zero-shot)}} \\
\cline{2-7} \cline{8-10}
\multicolumn{1}{l|}{} & \multicolumn{2}{c}{PEP(\%) $\downarrow$} & \multicolumn{1}{c}{\multirow{2}*{EPE(pixel) $\downarrow$}} & 
\multicolumn{1}{c}{\multirow{2}*{SSIM $\uparrow$}} & \multicolumn{1}{c}{\multirow{2}*{MSE $\downarrow$}} & \multicolumn{1}{c|}{\multirow{2}*{PSNR $\uparrow$}} &
\multicolumn{1}{c}{\multirow{2}*{\clr{SSIM* $\uparrow$}}} & \multicolumn{1}{c}{\multirow{2}*{\clr{MSE* $\downarrow$}}} & \multicolumn{1}{c}{\multirow{2}*{\clr{PSNR* $\uparrow$}}} \\

\cline{2-3}
\multicolumn{1}{l|}{} & \multicolumn{1}{c}{$\delta=0.5$} & \multicolumn{1}{c}{$\delta=1$} & \multicolumn{1}{c}{} & \multicolumn{1}{l}{} & \multicolumn{1}{c}{} & \multicolumn{1}{c|}{} & \multicolumn{1}{c}{} & \multicolumn{1}{c}{} & \multicolumn{1}{c}{} \\
\hline
\hline

\multicolumn{1}{l|}{PSMNet}              & \multicolumn{1}{c}{53.50}    & \multicolumn{1}{c}{22.00} & 2.57 & 0.82 & 152.2 & 28.3  & \clr{0.84} & \clr{129.0} & \clr{28.3} \\ 
\multicolumn{1}{l|}{PSMNet+synthetic}    & \multicolumn{1}{c}{\textbf{7.61}}    & \multicolumn{1}{c}{\textbf{3.59}} & \textbf{0.67} & \textbf{0.90} & \textbf{93.5} & \textbf{30.9} & \clr{\textbf{0.92}} & \clr{\textbf{64.7}} & \clr{\textbf{31.1}} \\ 
\hline
\multicolumn{1}{l|}{AANet}               & \multicolumn{1}{c}{37.60}    & \multicolumn{1}{c}{23.7} & 4.79 & 0.85 & 118.2 & 29.7  & \clr{0.77} & \clr{198.7} & \clr{25.9} \\ 
\multicolumn{1}{l|}{AANet+synthetic}     & \multicolumn{1}{c}{\textbf{9.36}}    & \multicolumn{1}{c}{\textbf{5.30}} & \textbf{0.61} & \textbf{0.91} & \textbf{93.4} & \textbf{30.8} & \clr{\textbf{0.93}} & \clr{\textbf{67.7}} & \clr{\textbf{30.7}} \\ 
\hline
\multicolumn{1}{l|}{LacGwc}              & \multicolumn{1}{c}{15.00}    & \multicolumn{1}{c}{8.17} & 1.35 & 0.89 & 113.4 & 30.4 & \clr{0.92} & \clr{75.8} & \clr{30.3} \\ 
\multicolumn{1}{l|}{LacGwc+synthetic}    & \multicolumn{1}{c}{\textbf{6.11}}    & \multicolumn{1}{c}{\textbf{2.92}} & \textbf{0.55} & \textbf{0.90} & \textbf{96.0} & \textbf{30.7} & \clr{\textbf{0.93}} & \clr{\textbf{63.3}} & \clr{\textbf{31.1}} \\ 
\hline
\multicolumn{1}{l|}{IGEV}                & \multicolumn{1}{c}{8.14}    & \multicolumn{1}{c}{4.99} & 0.83 & 0.90 & 102.9 & 30.8 & \clr{0.93} & \clr{64.2} & \clr{31.1} \\ 
\multicolumn{1}{l|}{IGEV+synthetic}      & \multicolumn{1}{c}{\textbf{3.17}}    & \multicolumn{1}{c}{\textbf{1.59}} & \textbf{0.23} & \textbf{0.91} & \textbf{92.5} & \textbf{30.9} & \clr{\textbf{0.93}} & \clr{\textbf{60.7}} & \clr{\textbf{31.3}} \\ 
\hline
\multicolumn{1}{l|}{ViTAStereo}          & \multicolumn{1}{c}{7.61}    & \multicolumn{1}{c}{5.08} & 1.31 & 0.90 & 124.4 & 29.6 & \clr{0.92} & \clr{79.6} & \clr{29.9} \\ 
\multicolumn{1}{l|}{ViTAStereo+synthetic}& \multicolumn{1}{c}{\textbf{3.14}}    & \multicolumn{1}{c}{\textbf{1.61}} & \textbf{0.22} & \textbf{0.91} & \textbf{91.6} & \textbf{30.9} & \clr{\textbf{0.93}} & \clr{\textbf{60.0}} & \clr{\textbf{31.3}} \\ 

\bottomrule
\end{tabular}
}
}
\end{table*}


\begin{figure*}[t!]
    \centering
    \includegraphics[width=0.995\textwidth]{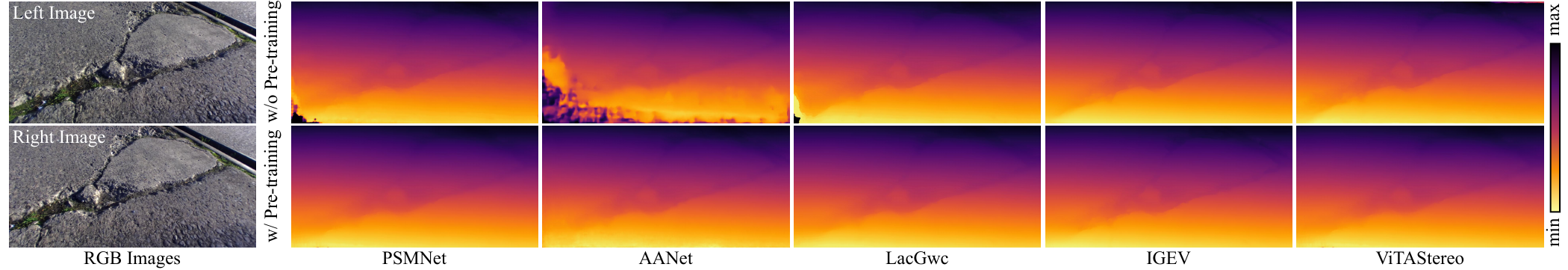}
    \caption{ 
    \clr{Depth visualizations for stereo matching on real-world images, comparing models with and without pre-training on synthetic road defect data.} 
    }
    \label{fig.stereo_viz}
\end{figure*}

\subsection{\cll{Decision-Making} Tasks}
Two types of experiments are conducted: one for 2D path planning, where road defects are avoidable, and the other for speed control, where the vehicle must traverse defects.

\subsubsection{Path Planning}  \label{subsec:path_planning}
In scenarios with ample road space and minimal opposing traffic, avoiding defects by following an alternative path is generally the preferred strategy. 
Unlike traditional obstacle constraints, road defects primarily impact the wheels rather than the vehicle body, allowing the vehicle to either bypass or glide over the defects. 
As shown in Fig. \ref{fig.pnc_tasks}(a), multiple defects are distributed along the road. The drivable area and defected regions are modeled as grid maps with a resolution of 0.2m, and the vehicle model assumes a width of 2.0m and a rear-to-front wheel distance of 3.0m. 
\cll{Four widely used planning algorithms, A*, RRT* , Lattice, and Hybrid A*, are compared.}
Evaluation metrics \cll{follow those} from classical obstacle avoidance tasks. Path deviation quantifies the length difference between the planned path and the shortest possible path, typically the road's midline. Path smoothness is calculated by averaging the angle changes along the trajectory, while obstacle clearance measures the mean distance between the vehicle's wheels and the nearest road defects.
Quantitative and qualitative results of the generated paths are presented in Table \ref{tab:path_planning} and Fig. \ref{fig.pnc_tasks}(a). Notably, A* and RRT* do not account for vehicle kinodynamic constraints, resulting in relatively poor path smoothness. However, their discrete path points remain topologically reasonable. The paths generated by Hybrid A* and Lattice glide over defects $P_0$ and $P_1$, which is justified in the given context\clb{, respectively}.
Lattice generates the shortest and smoothest path, with a path deviation of 0.87 and a smoothness value of 0.047. However, because its grids and curves are pre-sampled, obstacles may not always be perfectly avoided. 
In contrast, Hybrid A* expands nodes iteratively at each step, achieving an obstacle clearance of 3.74, which ensures better safety distances. The trade-off, however, is that heuristic terms must be manually designed to suit specific scenarios.

\subsubsection{Speed Control}  \label{subsec:speed_control}
\begin{figure}[!htbp]
    \centering
    \includegraphics[width=0.47\textwidth]{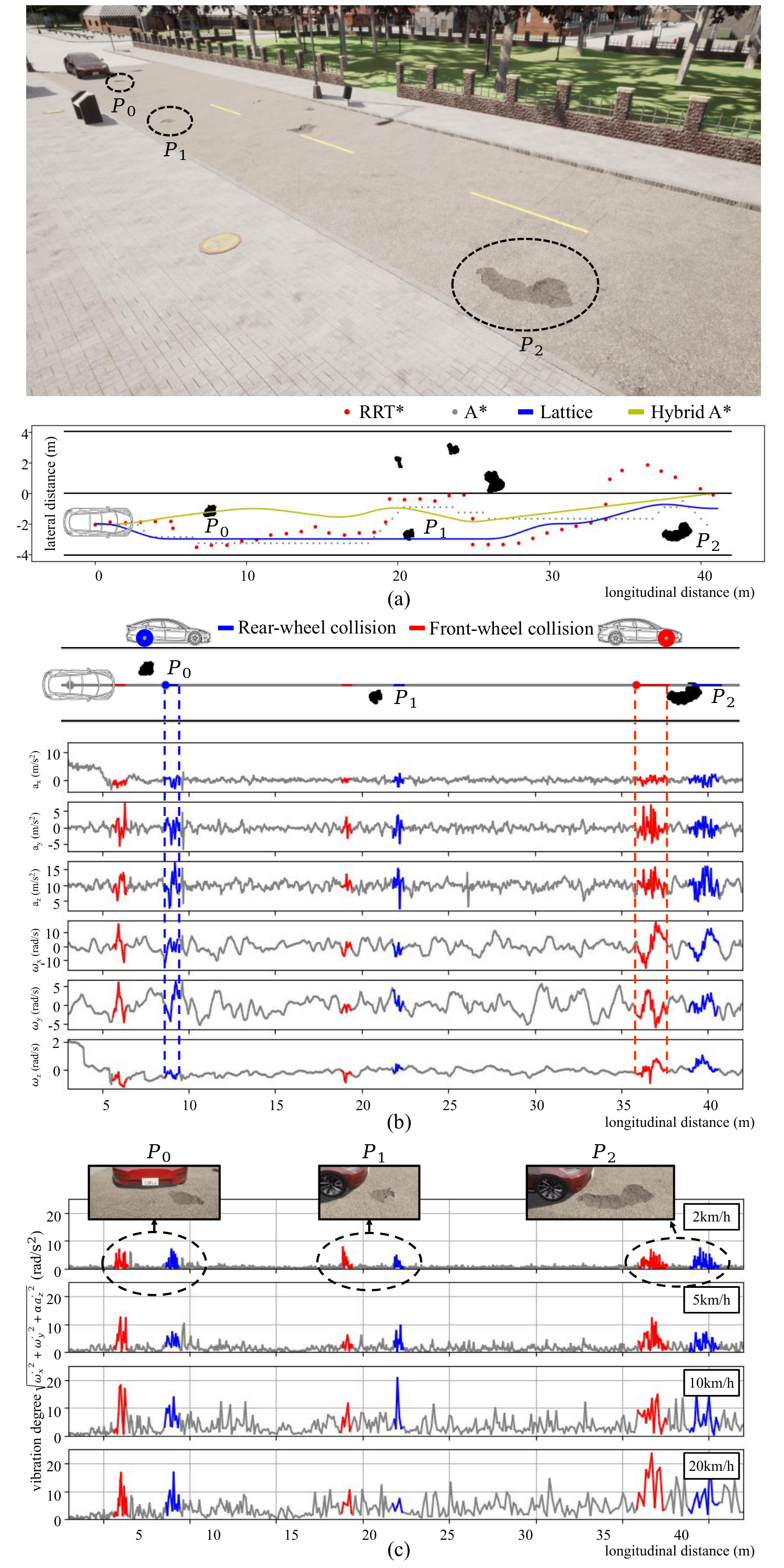}
    \caption{Experiments on planning and control tasks using our UDT system. (a) Path planning for road defect avoidance, focusing on tire-ground collision constraints instead of vehicle body. (b) \clb{I}nertial measurements recorded while traversing road defects at a constant speed of 5 km/h. (c) Vibration degree evaluated over road defects with different traversing speeds.
    }
    \label{fig.pnc_tasks}
\end{figure}

In scenarios where lane changes pose a high risk, the vehicle must traverse road defects directly. In such cases, longitudinal speed becomes the only controllable variable to minimize vibrations.
We customize these scenarios using our UDT system, where real-world road surface models are cropped into 40-meter sections, and integrated with randomly distributed road defects. With the aid of simulators, road defects along the wheel trajectories can be precisely located, enabling informed speed control strategies.
To measure vehicle vibrations caused by defected road surfaces,
collision detection is detected at the level of individual triangular faces between the road and tires during physical simulation.
As shown in Fig. \ref{fig.pnc_tasks}(b), six-axis IMU data is recorded as the vehicle drives at a constant speed.
During the process of a wheel falling into and exiting a road defect, it can be observed that the most affected parameters are vertical acceleration $ a_z $, rotational velocity in roll $\omega_x$ and pitch $\omega_y$. 
Subsequently, the vibration degree is defined as $ g = \sqrt{{\omega_x'}^2 + {\omega_y'}^2 + \alpha {a_z'}^2 }$, where $\alpha$ is set to 0.1 in this scenario.
Quantitative results for traversing three road defects at different speeds are presented in Fig. \ref{fig.pnc_tasks}(c).
The results indicate that, for large and severe defects such as $P_2$, slowing down is the most effective strategy for driving comfort.
Interestingly, for small and shallow defects, such as $P_1$, increasing speed can also reduce vibrations. Although initially counterintuitive, this observation aligns with practical experience upon closer analysis.

\begin{table}[t!]
\caption{Comparison of path planning algorithms to avoid road defects}
\label{tab:path_planning} 
\centering
\settablefont
\setlength{\tabcolsep}{4mm}{
\begin{tabular}{c c c c}
\toprule
\multicolumn{1}{c|}{\multirow{2}*{Planner}} & \multicolumn{1}{c}{path} & \multicolumn{1}{c}{path smooth-} & \multicolumn{1}{c}{obstacle}\\
\multicolumn{1}{c|}{} & \multicolumn{1}{c}{dev.($\%$)} & \multicolumn{1}{c}{ness(rad/m)} & \multicolumn{1}{c}{ clearance(m)}\\
\hline
\hline
\multicolumn{1}{c|}{A*}          & \multicolumn{1}{c}{7.46}    & \multicolumn{1}{c}{0.294} & 3.69  \\ 
\hline
\multicolumn{1}{c|}{RRT*}        & \multicolumn{1}{c}{21.2}    & \multicolumn{1}{c}{0.402} & 3.72  \\ 
\hline
\multicolumn{1}{c|}{Lattice}     & \multicolumn{1}{c}{\textbf{0.87}}    & \multicolumn{1}{c}{\textbf{0.047}} & 2.62  \\ 
\hline
\multicolumn{1}{c|}{Hybrid A*}   & \multicolumn{1}{c}{1.05}    & \multicolumn{1}{c}{0.054} & \textbf{3.74}  \\ 
\bottomrule
\end{tabular}
}
\end{table}


\clr{\section{Discussion and Future Work}
\label{sec.discussion}
}
Despite these advancements, several challenges remain. 
\clv{
First, the defect taxonomy is not yet comprehensive. The current library primarily models geometric surface deformations such as potholes and depressions, whereas finer-scale surface distresses, including cracking, raveling, and bleeding, are not yet reconstructed. Achieving a more complete taxonomy will require larger-scale data acquisition and systematic defect surveying in future deployments.
}
Second, loss of geometric and texture fidelity is inevitable during the reconstruction process, leading to domain discrepancy between reconstructed virtual environments and real-world observations. 
As a result, synthetic data is primarily limited to pre-training and must be supplemented with real-world images for fine-tuning. 
\clr{
Future work could explore domain adaptation techniques and advanced rendering methods to reduce the sim-to-real gap. Another promising direction is to adopt novel view synthesis (NVS) approaches, which learn neural scene representations directly from multi-view imagery without explicit geometric reconstruction. Unlike traditional geometry pipelines, NVS has the potential to mitigate fidelity loss during reconstruction and deliver more photorealistic renderings for downstream tasks.
}
\clr{
Another limitation concerns deployment.
Due to constraints of our single prototype and the campus-scale environment, we have not yet explored large-scale, in-the-wild deployment scenarios that are essential for real-world road inspection.
In the future, the generated synthetic data can be exploited for model fine-tuning and knowledge distillation, thereby enabling the development of lightweight models capable of real-time operation in large-scale road monitoring applications.
}

\clr{\section{Conclusion}
\label{sec.conclusion}
}
\cll{This letter presents a UDT system that exploits reality–virtual interconnections of road entities for intelligent road inspection.
The system captures data from defective roads using a custom sensor setup, constructs detailed defect and surface models through a hierarchical road model creator, and integrates them into simulation scenes via a digital road twin generator to recover simplified planar road assets.}
The proposed system provides unlimited, multi-modal, and well-annotated data for training road inspection networks, as well as a physical simulation environment for testing autonomous driving tasks under defected road conditions.
\cll{Experiments show that pre-training semantic segmentation networks with our synthetic data substantially boosts performance, alleviating the shortage of annotated real-world defect data.
The precise modeling of road defects enables both trajectory planning to bypass defects and speed control to traverse them smoothly, balancing safety and comfort.
Finally, we discuss current limitations in addressing the sim-to-real gap and challenges in in-the-wild deployment, outlining future directions for improved realism and large-scale applications.}

\normalem

\bibliographystyle{IEEEtran} 
\bibliography{egbib}

\end{document}